%% file: main.tex
\documentclass[10pt,twocolumn,letterpaper]{article}

\usepackage{cvpr}              

\usepackage{multirow}
\usepackage{array}
\usepackage{bbding}
\usepackage{amssymb}  
\usepackage[accsupp]{axessibility}
\usepackage{bbding}

\input{preamble}
\definecolor{cvprblue}{rgb}{0.21,0.49,0.74}
\usepackage[pagebackref,breaklinks,colorlinks,allcolors=cvprblue]{hyperref}

\def\paperID{27486} 
\def\confName{CVPR}
\def\confYear{2026}

\title{Perceive and Calibrate: Analyzing and Enhancing Robustness of Medical Multi-Modal Large Language Models}

\author{
Dunyuan XU\textsuperscript{\rm 1$^{*}$}, Xikai Yang\textsuperscript{\rm 1$^{*,\dagger}$}, Yaoqian Li\textsuperscript{\rm 1}, Juzheng Miao\textsuperscript{\rm 1}, Jinpeng Li\textsuperscript{\rm 1}~\textsuperscript{\Envelope}, Pheng-Ann Heng\textsuperscript{\rm 1,2}
\vspace{0.2cm}\\
\textsuperscript{\rm 1}Department of Computer Science and Engineering, CUHK, Hong Kong, China;\\ 
\textsuperscript{\rm 2}Institute of Medical Intelligence and XR, CUHK, Hong Kong, China\\
$^{*}$ Equal contribution, $^{\dagger}$ Project lead, \Envelope 
 Corresponding author \\
jpli21@cse.cuhk.edu.hk
}

\begin{document}

\twocolumn[
{%
\renewcommand\twocolumn[1][]{#1}
\maketitle
\begin{center}
\centering
\begin{minipage}[t]{\linewidth}
\includegraphics[width=\textwidth]{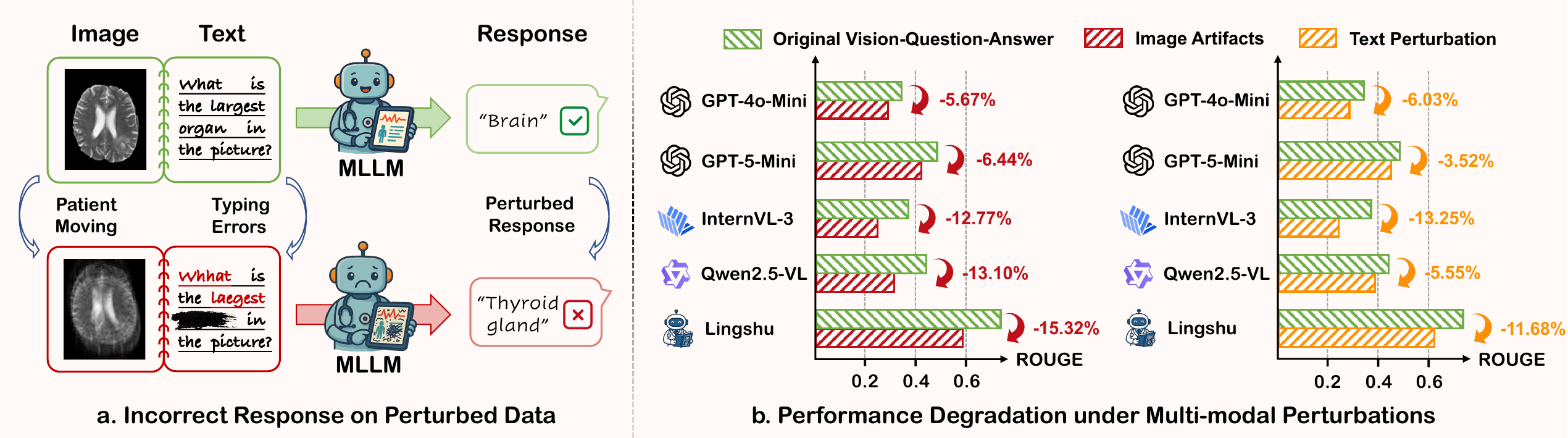}
\vspace{-0.6cm}
{\captionsetup{hypcap=false}  
\captionof{figure}{
Analysis of MLLM sensitivity to different input perturbations: (a) MLLM produces incorrect responses when inputs contain noise. (b) Comparison between original and perturbed VQA pairs, which demonstrates the performance deterioration of leading MLLMs under visual and textual perturbations (left: CT sparse view artifact, 
right: character level typographical errors).
}
\label{fig:teaser}}
\end{minipage}
\end{center}
}]

\maketitle

\input{sec/0_abstract}    
\input{sec/1_intro}

\input{sec/2_relate}
\input{sec/3_benchmark}

\input{sec/4_method}

\input{sec/5_exp}

\input{sec/6_con}

{
    \small
    \bibliographystyle{ieeenat_fullname}
    \bibliography{main}
}


\clearpage
\input{sec/X_suppl}

\end{document}

%% file: sec/0_abstract.tex
\begin{abstract}

Medical Multi-modal Large Language Models (MLLMs) have shown promising clinical performance.
However, their sensitivity to real-world input perturbations, such as imaging artifacts and textual errors, critically undermines their clinical applicability. 
Systematic analysis of such noise impact on medical MLLMs remains largely unexplored.
Furthermore, while several works have investigated the MLLMs' robustness in general domains, they primarily focus on text modality and rely on costly fine-tuning.
They are inadequate to address the complex noise patterns and fulfill the strict safety standards in medicine.
To bridge this gap, this work systematically analyzes the impact of various perturbations on medical MLLMs across both visual and textual modalities.
Building on our findings, we introduce a training-free Inherent-enhanced Multi-modal Calibration (IMC) framework that leverages MLLMs' inherent denoising capabilities following the perceive-and-calibrate principle for cross-modal robustness enhancement.
For the visual modality, we propose a Perturbation-aware Denoising Calibration (PDC) which leverages MLLMs' own vision encoder to identify noise patterns and perform prototype-guided feature calibration.
For text denoising, we design a Self-instantiated Multi-agent System (SMS) that exploits the MLLMs' self-assessment capabilities to refine noisy text through a cooperative hierarchy of agents.
We construct a benchmark containing 11 types of noise across both image and text modalities on 2 datasets.
Experimental results demonstrate our method achieves the state-of-the-art performance across multiple modalities, showing potential to enhance MLLMs' robustness in real clinical scenarios.

\end{abstract}

%% file: sec/1_intro.tex
\section{Introduction}
\label{sec:intro}

Medical Multi-modal Large Language Models (MLLMs) have achieved remarkable progress due to their superior capabilities across various tasks~\cite{ma2025medla,xu2024medvilam,xiao2025comprehensive} and potential to alleviate clinical workload \cite{zhang2024potential}.
However, their robustness in handling input noise remains far from optimal, as minor corruptions in the inputs can lead to dramatic changes in responses and result in erroneous outputs \cite{villani2025robust} (as shown in Figure \ref{fig:teaser}(a)).
Moreover, compared to the general domain, medical data commonly involve more diverse noise that is difficult to detect, such as equipment deterioration \cite{kelvin2024impact}, patient motion artifacts \cite{kavitha2023noise}, or human operational errors \cite{owusu2021factors}, leading to significant model performance decrease.
Several studies have highlighted the impact of prompt noise on the reliability of LLMs \cite{gu2023robustness,lu2024mitigating,agrawal2025enhancing,razavi2025benchmarking}.
Existing solutions include layer editing approaches that update targeted parameters \cite{liu2025improving} and adversarial training methods that expose models to constructed perturbed examples \cite{bai2025enhancing}. 
However, these investigations predominantly focus on text-modality under general domains and require resource-intensive fine-tuning processes, making them unsuitable for clinical practice.
In this work, we tackle this critical challenge from two perspectives: 1) Analyzing the impact of prevalent noise types in medical images and questions on medical MLLMs performance; 2) Leveraging these insights to develop a training-free framework for enhancing medical MLLMs robustness.

To systematically quantify the model vulnerability, we propose, to the best of our knowledge, \textit{the first comprehensive benchmark, RobustMed-Bench}, which analyzes the sensitivity of MLLMs to different perturbed inputs under medical scenarios.
Our RobustMed-Bench simulates diverse noise across both visual and textual modalities, generating original-noisy data pairs that provide clear insights into model performance variations under distinct noisy conditions.
For image modalities, we select six types of image noise commonly arising in three major types of medical imaging, including MRI, CT, and X-ray.
For the text modality, we analyze typical patterns in how humans ask medical questions and simulate five prevalent types of noise at both the character level and the sentence level \cite{gu2023robustness}.
Extensive experiments on our RobustMed-Bench reveal that existing leading MLLMs show notable performance deterioration under both visual and textual perturbations (as recorded in Figure \ref{fig:teaser}(b)), highlighting the urgent need for robustness improvements to enable safe clinical integration.

To counter this fragility, we propose a training-free Inherent-enhanced Multi-modal Calibration (IMC) framework based on the principle of ``perceive-and-calibrate".
Our IMC is built based on the insight that MLLMs can inherently perceive and correct multi-modal noise, whose abilities can be unlocked without any finetuning or external modules.
Our experimental results reveal that MLLMs' vision encoders can effectively extract rich latent information for precise fine-grained classification of different states, distinguishing both normal cases across modalities and specific abnormal patterns up to specific noise types.
Based on this finding, we design an effective Perturbation-aware Denoising Calibration (PDC) that utilizes the MLLMs' built-in vision encoders to compute embedding prototypes and feature gaps between original and noisy images, employing these results for perturbation classification and fine-grained noise calibration.
Meanwhile, we observe that MLLMs demonstrate a capability to partially identify and correct textual errors when properly prompted.
Inspired by this, we propose a Self-instantiated Multi-agent System (SMS) for text denoising.
The system first parallelly coordinates diverse self-initiated agents for perceiving and removing textual noise.
Then it aggregates and refines these denoised outputs with visual information, feeding the result back for the next parallel denoising loop.

Our main contributions are summarized as follows:

\begin{itemize}
    \item We construct RobustMed-Bench to systematically analyze noise effects on medical MLLMs through a noise simulation pipeline that generates common medical noise types across both image and text modalities.
    \item We empirically identify that multi-modal perturbations result in substantial performance degradation among existing leading MLLMs through extensive experiments.
    \item We develop a training-free multi-modal IMC framework consisting of PDC and SMS that both utilize perceive-and-calibrate processes for image and text denoising.
    \item The experimental results show that our method greatly improves model robustness against diverse noise, advancing the clinical applicability of medical MLLMs.
\end{itemize}

%% file: sec/2_relate.tex
\section{Related Work}
\label{sec:relate}

\subsection{Noise in Medical Multi-Modal Data}
Medical data from real-world clinical settings is susceptible to highly complex and diverse types of noise across both image and textual modalities \cite{cheng2025med,kumar2024multimodality,yang2024computational}.
Medical image noise stems from multiple corruption sources, such as instrumental artifacts \cite{huang2025deep}, environmental factors \cite{shaik2024classification}, modality distortions \cite{hemalatha2025masked}, and registration errors \cite{alshmrani2023hyper}.
Meanwhile, textual noise in medical data, such as non-standardized language \cite{meyer2024impatient} and transcription errors \cite{weerasinghe2024real}, complicates the extraction of accurate semantics.
Several recent works have identified the challenge in handling multi-modal noise and tried to tackle it through tailored denoising strategies, such as training additional deep learning-based denoising models \cite{soy2025medical,li2021novel,atal2023optimal,rahman2023task}, employing self-supervised denoising \cite{zhou2024neighboring,xu2021deformed2self,jeon2025unsupervised}, and leveraging data enhancement strategies \cite{lerch2024dreamon,kebaili2023deep}.
However, these studies mainly focus on removing noise in conventional deep learning frameworks.
Consequently, a significant gap persists in understanding how multi-modal medical data noise impacts MLLMs, which is vital for the reliability of these advanced models deployed in clinical scenarios.
Therefore, in this work, we conduct a systematic investigation and propose solutions for alleviating perturbation impacts on medical MLLMs.

\subsection{Sensitivity of LLMs to Perturbations}
Prior research has examined the robustness of LLMs to text noise \cite{gu2023robustness,lu2024mitigating,agrawal2025enhancing,razavi2025benchmarking}, with several studies creating benchmarks for robustness evaluation, such as Lmentry \cite{efrat2023lmentry}, PromptEval \cite{polo2024efficient}, and PromptBench \cite{zhu2023promptbench}.
Based on these studies, researchers have attempted to enhance model robustness through additional fine-tuning \cite{zhao2024improving,yi2024safety,wang2024enhancing}. 
However, such computationally expensive approaches are impractical for models that are already deployed.
Alternative researches have explored training-free methodologies by employing randomized smoothing \cite{he2025certifying,hu2025enhancing}. 
Nevertheless, these methods need extensive noisy samples to achieve reliable smooth distribution and inject random noise to the intermediate states.
This data-intensive requirement makes them unsuitable for complex multi-modal data.
Although Liu et al. \cite{liu2024reducing} have designed a multi-modal framework for reducing hallucinations in MLLMs, it is not applicable to some types of textual perturbations that introduce out-of-vocabulary tokens beyond the tokenizer's encoding capacity.
Therefore, there is an urgent need to design a comprehensive framework for enhancing model robustness against diverse multi-modal noise types without additional fine-tuning or external models.

%% file: sec/3_benchmark.tex
\section{RobustMed-Bench}
\label{sec:bench}

\begin{figure*}[h]
  \centering
   \includegraphics[width=0.99\linewidth]{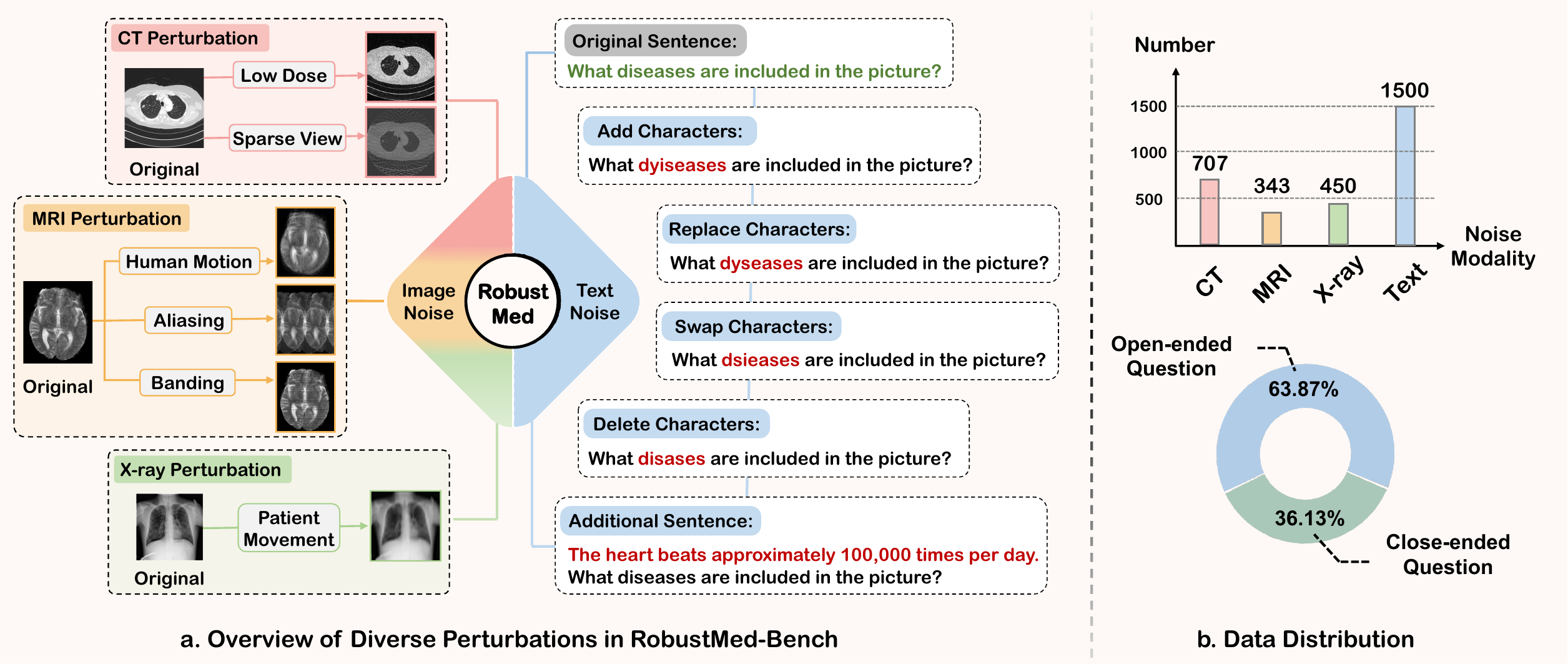}
  \caption{Overview of SLAKE-based RobustMed-Bench dataset composition: (a) Visualization of noise types included in our benchmark. For images, CT contains low dose and sparse view noise, MRI includes human motion, aliasing and banding corruptions, and X-ray with patient movement artifacts. For text, we incorporate four common typographical errors: add/delete/swap/replace characters in words, and additional unrelated sentences noise; (b) Statistical distribution of modalities and question categories of the SLAKE-based benchmark.}
  \label{fig:dataset}
\end{figure*}

To systematically investigate the robustness challenges faced by medical MLLMs in real-world clinical scenarios, we introduce RobustMed-Bench, a comprehensive evaluation benchmark designed to assess model sensitivity to authentic medical data noise.
The construction of our benchmark is detailed below, with illustrative noisy cases provided in Figure \ref{fig:dataset}(a).

\subsection{Image noise}
For image noise construction, we consider three prevalent medical imaging modalities in clinical practice (CT, MRI and X-ray), simulating distinct noise types that are aligned with real clinical scenarios for each modality.
\begin{itemize}
    \item Noise in CT imaging primarily arises from clinical demands to minimize radiation exposure and enhance scanning efficiency, resulting in \textit{sparse view} \cite{lee2018deep} and \textit{low dose} \cite{moen2021low} noise. 
    Sparse view generates regular stripe artifacts degrading spatial resolution, whereas low dose induces randomly distributed granular noise that impairs contrast resolution. Both types are extremely common in practice.
    
    \item MRI noise mainly arises from the longer scanning time required compared to other medical imaging techniques and its inherent hardware constraints.
    \textit{Patient motion} generates ghosting artifacts that blur anatomical details \cite{havsteen2017movement}, radio frequency interference creates \textit{banding} noise with regular stripe patterns \cite{defazio2020mri}, and insufficient field of view relative to imaged anatomy produces \textit{aliasing} noise with overlapping structure artifacts \cite{zhao2019applications}.
        
    \item Unlike other imaging modalities, X-ray acquisition occurs instantaneously, making it highly sensitive to even minimal movements \cite{karim2023portable}. 
    \textit{Patient movement} or heavy respiratory motion during the scanning process can produce streaking artifacts and reduced contrast resolution.
    
\end{itemize}    

In RobustMed-Bench, we incorporate different severity levels of these noise into original images from SLAKE \cite{liu2021slake} and OmniMed \cite{hu2024omnimedvqa}, due to their rich diversity in medical imaging modalities and question types.
In the main paper, we focus on the SLAKE-based dataset, while details of the OmniMed-based dataset are presented in the supplementary.
As shown in Figure \ref{fig:dataset}(b), our SLAKE-based benchmark contains 707/343/450 samples for CT/MRI/X-ray modalities respectively, with open-ended and closed-ended questions comprising 64\% and 36\% of the dataset.

\subsection{Text noise}
We inject various common typographical errors into question sentences to simulate real-world text input scenarios at both character and sentence levels.
For character level noise, we implement four types: 1) random character insertion within words, where arbitrary characters are inserted at random positions to mimic typing mistakes; 2) random character deletion from words, simulating cases where users accidentally miss keystrokes; 3) random character transposition within words, reflecting common finger placement errors where adjacent characters are swapped; 4) random character substitution within words, representing scenarios where users hit incorrect keys due to adjacent key confusion on standard keyboards.
For sentence level noise, we recognize that clinicians may provide extraneous information beyond the main question when interacting with MLLMs, such as background context.
Therefore, we also incorporate extra sentence noise by adding semantically unrelated sentences alongside the primary medical question.

\begin{table}[h]
\centering
\fontsize{8.5pt}{8.5pt}\selectfont
\addtocounter{table}{0}
\renewcommand{\arraystretch}{1}
\setlength{\tabcolsep}{1.1mm}
{
\caption{Performance degradation of MLLMs under CT sparse view artifacts and mixed character level text noise (including randomly delete/add/swap/replace). Subscripts show degradation.}
\label{tab:identify_problem}
\begin{tabular}{>{\centering\arraybackslash}l|c|cc}
\toprule
MLLMs & Noise & ACC($\uparrow$) & ROUGE($\uparrow$) \\
\midrule
\multirow{3}{*}{\raisebox{-0ex}{GPT-5-Mini \cite{openai2025gpt5}}} 
& Original & 79.64 & 49.23 \\
& CT Sparse View & $\text{78.21}_{-1.43}$ & $\text{42.78}_{-6.45}$ \\
& Character Noise & $\text{77.14}_{-2.50}$ & $\text{45.71}_{-3.52}$ \\

\midrule
\multirow{3}{*}{\raisebox{-0ex}{GPT-4o-Mini \cite{hurst2024gpt}}} 
& Original & 70.71 & 34.78 \\
& CT Sparse View & $\text{65.71}_{-5.00}$ & $\text{29.11}_{-5.67}$ \\
& Character Noise & $\text{66.79}_{-3.92}$ & $\text{28.75}_{-6.03}$ \\

\midrule
\multirow{3}{*}{\raisebox{-0ex}{InternVL-3-9B \cite{zhu2025internvl3}}} 
& Original & 75.71 & 37.67 \\
& CT Sparse View & $\text{64.29}_{-11.42}$ & $\text{24.90}_{-12.77}$ \\
& Character Noise & $\text{59.29}_{-16.42}$ & $\text{24.42}_{-13.25}$ \\

\midrule
\multirow{3}{*}{\raisebox{-0ex}{Qwen2.5-VL-7B \cite{bai2025qwen2}}} 
& Original & 73.93 & 44.98 \\
& CT Sparse View & $\text{60.71}_{-13.22}$ & $\text{31.89}_{-13.09}$ \\
& Character Noise & $\text{66.79}_{-7.14}$ & $\text{39.43}_{-5.55}$ \\

\midrule
\multirow{3}{*}{\raisebox{-0ex}{Lingshu-7B \cite{xu2025lingshu}}} 
& Original & 78.57 & 75.10 \\
& CT Sparse View & $\text{77.14}_{-1.43}$ & $\text{59.78}_{-15.32}$ \\
& Character Noise & $\text{69.29}_{-9.28}$ & $\text{63.42}_{-11.68}$ \\

\bottomrule
\end{tabular}}
\end{table}

\subsection{Sensitivity of MLLMs to noise}
This subsection explores the robustness of multiple state-of-the-art MLLMs against the previously described noise using RobustMed-Bench.
We report accuracy and ROUGE-1 scores for closed-ended and open-ended questions respectively, using CT sparse view for visual noise and a random combination of four character noise types for textual disturbances.
To maintain clear presentation, complete results for all noise variations are provided in the supplementary material.
As shown in Table \ref{tab:identify_problem}, all tested MLLMs exhibit notable performance degradations under noisy conditions.
Closed-source MLLMs (GPT-5-Mini\cite{openai2025gpt5} and GPT-4o-Mini\cite{hurst2024gpt}) experience approximately 7\% and 5\% ROUGE score reductions respectively under image noise, and 4\% and 6\% performance drops under text noise compared to original data.
The degradation is more severe for open-source MLLMs, including general-domain models (InternVL-3-9B\cite{zhu2025internvl3}, Qwen2.5-VL-7B\cite{bai2025qwen2}) and medical-specific model (Linshu-7B\cite{xu2025lingshu}), where performance losses on open-ended questions can reach approximately 13\%-15\% under image noise.
The situation for closed-ended questions is also unsatisfactory.
GPT-5-Mini exhibits the highest robustness among all tested MLLMs, showing merely 1.43\% and 2.50\% accuracy drops under image and text noise, respectively. 
In contrast, InternVL-3-9B, Qwen2.5-VL-7B suffer more than 10\% accuracy loss on image corruption and 16\% and 7\% respectively on text noise.
Although Lingshu-7B was trained on SLAKE data and shows better robustness on closed-ended questions than other open-source MLLMs, it still struggles with input noise with significant performance drops on open-ended questions.
To make a brief summary, current MLLMs demonstrate inadequate robustness against noise in medical datasets, resulting in substantial performance deterioration that presents a major obstacle to practical medical deployment, thus highlighting the urgent need for developing an efficient denoising framework.

%% file: sec/4_method.tex
\section{Method}
\label{sec:Method}

\begin{figure*}[h]
  \centering
   \includegraphics[width=1\linewidth]{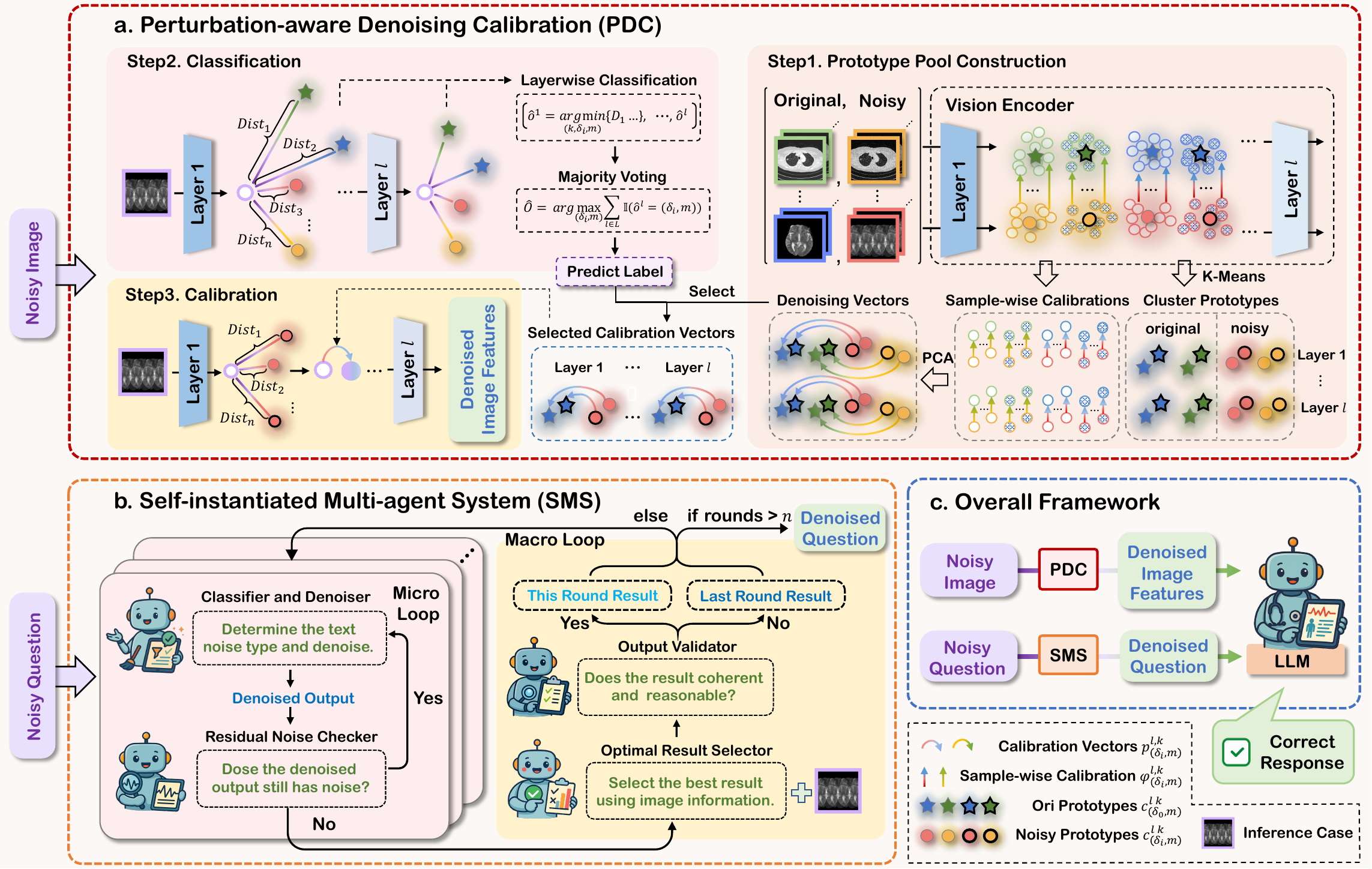}
  \caption{Architecture of our Inherent-enhanced Multi-modal Calibration (IMC) framework, which follows the perceive-and-calibrate process for both image and text modalities. (a) Perturbation-aware Denoising Calibration (PDC) enables image denoising through prototype-guided classification, then applies PCA-based calibration vectors to rectify vision features across all vision encoder layers; (b) Self-instantiated Multi-agent System (SMS) enables text denoising through hierarchical multi-agent coordination with micro and macro iterations that simulate human iterative editing processes; (c) Our framework integrates PDC and SMS for unified multi-modal denoising.
  }
  \label{fig:pipeline}
\end{figure*}

\subsection{Preliminary}
\label{method:preliminary}
In this task, we define several types of noise 
that can be applied to original Vision($v$)-Quesion($q$)-Answer($a$) (VQA) pairs $x =(v_{}, q_{}, a)$ by injecting them into either the visual or the textual part of the prompt, resulting in $(\tilde{v}, q, a)$ or $(v, \tilde{q}, a)$, respectively.
Our target is to build a comprehensive framework that enhances MLLM $\Theta$ to maintain consistent predictions under perturbations, formally expressed as $\Theta(\tilde{v}, q) \approx \Theta(v, \tilde{q}) \approx \Theta(v, q)$.
To improve the robustness of MLLMs while avoiding additional resource costs, we propose a training-free Inherent-enhanced Multi-modal Calibration (IMC) framework focusing on exploring only the model's inherent capabilities for multi-modal denoising.

\subsection{Perturbation-aware Denoising Calibration}
\label{method:PATIS}
We adhere to the ``perceive-and-calibrate" principle in designing the image denoising part for visual embedding calibration 
(Figure \ref{fig:pipeline}(a)).
Specifically, we first perceive the existence and type of noise by recruiting a prototype-based classifier across all vision encoder layers.
Afterward, a calibration vector is defined as the direction of principal embeddings from the noisy sample group toward the corresponding normal sample group.
Finally, the input noisy embedding is rectified by the calibration vector at each layer. 
The specific computation process is detailed as follows.

\textbf{Noise perception.} 
We achieve robust noise classification by constructing a prototype pool.
We first extract a group of embeddings $\{f^l_{(\delta_i, m)}\}$ from all samples belonging to modality $m \in M$ and noisy type $\delta_i \in \Delta$ at the $l$-th layer of the vision encoder.
Here, $M = \{\text{CT}, \text{MRI}, \text{X-ray}\}$ denotes three modalities considered in our experiment, and $\Delta = \{\delta_0, \delta_1, \dots, \delta_n\}$ represents all possible image states, where $\delta_0$ denotes the normal state and $\{\delta_1, \dots, \delta_n\}$ correspond to $n$ types of noise.
For each condition $(\delta_i, m)$, we generate a set of $K$ prototypes, $\{c^{l,k}_{(\delta_i, m)}\}_{k=1}^K$, by applying K-Means \cite{jain1988algorithms} clustering to the corresponding embeddings:
\begin{equation}
\{c^{l,k}_{(\delta_i, m)}\}_{k=1}^K = \textrm{KMeans}\big(\{f^l_{(\delta_i, m)}\}\big),
\label{eq:proto}
\end{equation}
where each prototype $c^{l,k}_{(\delta_i, m)}$ represents the cluster center of embeddings obtained from state $(\delta_i, m)$ and feature layer $l$.
During the inference stage, the image of the incoming sample $\hat{v}$ is fed into the vision encoder to obtain the layerwise embeddings $\{\hat{f}^l\}_{l=1}^L$.
For each layer, we select the nearest prototype $\hat{c}^{l,k}_{(\delta_i, m)}$ for $\hat{f}^l$ from all prototypes at the same layer, denoted as $\{\hat{c}^{l,k}_{(\delta_i, m)}\}_{k,\delta_i, m \in K\times \Delta \times M}$.
Thus, we obtain the classified result $\hat{o}^l=(\hat{\delta}^l_i, \hat{m}^l)$ at the $l$-th layer, including noise type and modality based on the identification of $\hat{c}^{l,k}_{(\delta_i, m)}$ by calculating the feature distance $\textrm{Dist}(,)$:
\begin{equation}
\hat{o}^l=(\hat{\delta}^l_i, \hat{m}^l) = \underset{k,\delta_i, m \in K\times \Delta \times M}{\textrm{arg}\min} { \textrm{Dist}(\hat{f}^l, c^{l,k}_{(\delta_i, m)}) }.
\label{eq:min_dist}
\end{equation}
Finally, we aggregate the predictions from all layers and apply majority voting to classify the image as the type that appears most frequently across layers:
\begin{equation} 
\hat{o}_{\hat{v}} = \underset{(\delta_i, m)}{\textrm{arg}\max} 
\sum_{l\in L}\mathbb{I}(\hat{o}^l =(\delta_i,m)),
\label{eq:majority_voting}
\end{equation}
where $\mathbb{I}(\cdot)$ is the indicator function that returns 1 if the condition is true and 0 otherwise.
The final classification result is denoted as $\underline{\underline{\hat{o}_{\hat{v}}=(\hat{\delta}_i, \hat{m})}}$.
If the predicted $\hat{\delta}_i=\delta_0$, we determine this is a normal image and process it directly. Otherwise, the image is forwarded to the calibration pipeline.

\textbf{Embedding calibration.}
After determining the noise type $\hat{\delta}_i$ and modality $\hat{m}$, we perform finer-grained selection at each layer to mitigate potential layer-wise misclassifications before majority voting. 
We select the optimal cluster $\hat{k}$ at each layer that is closest to the incoming embedding under the state $(\hat{\delta}_i, \hat{m})$ similar to Eq. \ref{eq:min_dist}.
Subsequently, the sample-wise calibration directions $\varphi_{(\hat{\delta}_i, \hat{m}),j}^{l,\hat{k}}$ are computed by subtracting the embeddings of noisy samples from the corresponding embeddings of normal samples:
\begin{equation}
\varphi_{(\hat{\delta}_i, \hat{m}),j}^{l,\hat{k}} = f^{l,\hat{k}}_{(\hat{\delta}_0, \hat{m}),j}- f^{l,\hat{k}}_{(\hat{\delta}_i, \hat{m}),j}, \quad \delta_i\neq \delta_0.
\label{eq:calibrate_direction}
\end{equation}
where $j \in \mathcal{G}_{(\hat{\delta}_i, \hat{m})}^{l,k}$ which denotes the set of sample indices associated with the $\hat{k}$-th cluster under the $(\hat{\delta}_i, \hat{m})$ state at the $l$-th layer.
Afterward, we derive the final calibration vector for the calibration direction group using the PCA algorithm:
\begin{equation} 
p_{(\hat{\delta}_i, \hat{m})}^{l,\hat{k}} = \text{PCA}\big(\{\varphi_{(\hat{\delta}_i, \hat{m}),j}^{l,\hat{k}}\}\big), \quad j \in \mathcal{G}_{(\hat{\delta}_i, \hat{m})}^{l,\hat{k}}.
\label{eq:carlib}
\end{equation}
This calibration vector can be interpreted as a perturbation-specific transformation in the latent space, guiding noisy embeddings toward the correct feature space.
During the embedding calibration process, the selected layer-wise calibration vectors are applied to rectify the embeddings of the noisy image at each vision encoder layer, with a calibration weight $\alpha$ controlling the modification severity: $\hat{f}^l = \hat{f}^l + \alpha \cdot p_{(\hat{\delta}_i, \hat{m})}^{l,\hat{k}}$.
The final denoised visual embedding will be served as the input for downstream LLM reasoning:
\begin{equation} 
\textrm{Response}=LLM(\hat{f}^L, \hat{q}).
\label{eq:response}
\end{equation}

\subsection{Self-instantiated Multi-agent System}
\label{method:HMA}
The text denoising module employs a self-instantiated multi-agent system consisting of two loops (Figure \ref{fig:pipeline}(b)). 
Specifically, the micro loop first perceives and removes textual noise. 
The macro loop then selects and refines outputs from multiple parallel micro loop iterations, feeding the refined result back as input for subsequent micro loops.

\textbf{Micro loop} incorporates an agent called the \textit{Classifier and Denoiser} which receives input sentences to identify noise types and performs corresponding noise removal operations. Subsequently, another agent called the \textit{Residual Noise Checker} determines whether the denoised results from previous step still contain noise, and if noise is detected, it invokes the \textit{Denoiser} again to perform additional denoising operations.
The micro loop terminates only when the \textit{Residual Noise Checker} confirms that the current result contains no more perturbations.
We perform this process $k$ times in parallel, yielding $k$ responses from the micro loops to broaden the range of denoising outcomes and thereby increase the probability of obtaining correct results.

\textbf{Macro loop} deploys a new agent named \textit{Optimal Result Selector}, which integrates the $k$ denoised results from the micro loop together with the input image and selects the result that is both noise-free and consistent with the image content.
To ensure the quality of the macro loop output, we introduce an \textit{Output Validator} that evaluates whether the result from previous \textit{Optimal Result Selector} represents a coherent and logically consistent sentence, thus mitigating potential confusion that may manifest in some MLLMs with limited language reasoning capabilities.
If the output is valid, we forward the updated sentence to the \textit{Denoiser} for the next micro loop iteration with a halved temperature parameter to enable finer-grained adjustments; otherwise, we forward the original input sentence from the current round to prevent the propagation of additional noise.
This hierarchical process ends when the macro loop reaches its predefined maximum number of rounds $n$, and the final denoised sentence is served as the new prompt for MLLM inference.

%% file: sec/5_exp.tex
\section{Experiment}
\label{sec:Exp}
\begin{table}[h]
\centering
\fontsize{8pt}{8pt}\selectfont
\addtocounter{table}{0}
\renewcommand{\arraystretch}{1.1}
\setlength{\tabcolsep}{1.2mm}
{
\caption{Performance comparison of MLLM robustness across six medical imaging artifacts, subscripts denote performance drops. \textbf{Bold} font denotes the best performance.}
\label{tab:img_pert}
\begin{tabular}{>{\centering\arraybackslash}c|cccc}
\toprule
Method & $\textrm{ACC($\uparrow$)}$ & $\textrm{ROUGE($\uparrow$)}$  & $\textrm{BLEU($\uparrow$)}$  & $\textrm{Recall($\uparrow$)}$ \\
\midrule
\midrule
\multicolumn{5}{c}{MRI Original VQA}\\
\midrule
Base Model & 77.05 & 45.95 & 8.07 & 45.23  \\
\midrule

\multicolumn{5}{c}{MRI Human Motion}\\
\midrule
Base Model & $\text{70.49}_{-6.56}$ & $\text{39.51}_{-6.44}$  & $\text{7.10}_{-0.97}$  & $\text{39.56}_{-5.67}$     \\
RoP \cite{mu2025robustness} & $\text{59.02}_{-18.03}$ & $\text{37.94}_{-8.01}$  & $\text{6.89}_{-1.18}$  & $\text{39.60}_{-5.63}$   \\
SD \cite{agrawal2025enhancing} & $\text{73.77}_{-3.28}$ & $\text{36.46}_{-9.49}$  & $\text{6.45}_{-1.62}$  & $\text{38.27}_{-6.96}$   \\
VTI \cite{liu2024reducing} & $\text{73.77}_{-3.28}$ & $\text{40.17}_{-5.78}$  & $\text{7.01}_{-1.06}$  & $\text{39.45}_{-5.78}$   \\
Ours & \textbf{$\text{75.41}_{-1.64}$} & \textbf{$\text{42.27}_{-3.68}$}  & \textbf{$\text{7.49}_{-0.58}$}  & \textbf{$\text{41.36}_{-3.87}$}   \\
\midrule

\multicolumn{5}{c}{MRI Aliasing}\\
\midrule
Base Model & $\text{54.10}_{-22.95}$ & $\text{18.36}_{-27.59}$  & $\text{3.22}_{-4.85}$  & $\text{17.92}_{-27.31}$  \\
RoP \cite{mu2025robustness} & $\text{54.10}_{-22.95}$ & $\text{31.12}_{-14.83}$  & $\text{5.61}_{-2.46}$  & $\text{31.38}_{-13.85}$   \\
SD \cite{agrawal2025enhancing} & $\text{55.74}_{-21.31}$ & $\text{28.01}_{-17.94}$  & $\text{4.98}_{-3.09}$  & $\text{28.94}_{-16.29}$    \\
VTI \cite{liu2024reducing} & $\text{55.74}_{-21.31}$ & $\text{20.88}_{-25.07}$  & $\text{3.47}_{-4.60}$  & $\text{19.01}_{-26.22}$    \\
Ours & \textbf{$\text{60.26}_{-16.79}$} & \textbf{$\text{31.64}_{-14.31}$}  & \textbf{$\text{5.65}_{-2.42}$}  & \textbf{$\text{31.61}_{-13.62}$}  \\
\midrule

\multicolumn{5}{c}{MRI Banding}\\
\midrule
Base Model & $\text{75.41}_{-1.64}$ & $\text{43.86}_{-2.09}$  & $\text{7.65}_{-0.42}$  & $\text{42.51}_{-2.72}$    \\
RoP \cite{mu2025robustness} & $\text{70.49}_{-6.56}$ & $\text{41.21}_{-4.74}$  & $\text{7.23}_{-0.84}$  & $\text{40.67}_{-4.56}$   \\
SD \cite{agrawal2025enhancing} & $\text{73.77}_{-3.28}$ & $\text{41.27}_{-4.68}$  & $\text{7.07}_{-1.00}$  & $\text{41.33}_{-3.90}$    \\
VTI \cite{liu2024reducing} &  $\text{76.29}_{-0.76}$ & $\text{44.15}_{-1.80}$  & $\text{7.71}_{-0.36}$  & $\text{43.58}_{-1.65}$     \\
Ours & \textbf{$\text{77.00}_{-0.05}$} & \textbf{$\text{45.52}_{-0.43}$}  & \textbf{$\text{7.94}_{-0.13}$}  & \textbf{$\text{45.16}_{-0.07}$}    \\
\midrule
\midrule

\multicolumn{5}{c}{CT Original VQA}\\
\midrule
Base Model & 73.48 & 46.07 & 8.18 & 45.88  \\
\midrule

\multicolumn{5}{c}{CT Sparse View}\\
\midrule
Base Model & $\text{59.85}_{-13.63}$ & $\text{33.70}_{-12.37}$  & $\text{6.01}_{-2.17}$  & $\text{33.17}_{-12.71}$   \\
RoP \cite{mu2025robustness} & \textbf{$\text{63.64}_{-9.84}$} & $\text{31.61}_{-14.46}$  & $\text{5.63}_{-2.55}$  & $\text{31.32}_{-14.56}$   \\
SD \cite{agrawal2025enhancing} & $\text{62.12}_{-11.36}$ & $\text{30.64}_{-15.43}$  & $\text{5.66}_{-2.52}$  & $\text{35.69}_{-10.19}$    \\
VTI \cite{liu2024reducing} & $\text{56.06}_{-17.42}$ & $\text{32.41}_{-13.66}$  & $\text{5.87}_{-2.31}$  & $\text{32.93}_{-12.95}$   \\
Ours & $\text{63.14}_{-10.34}$ & \textbf{$\text{36.59}_{-9.48}$}  & \textbf{$\text{6.47}_{-1.71}$}  & \textbf{$\text{36.73}_{-9.15}$}   \\
\midrule

\multicolumn{5}{c}{CT Low Dose}\\
\midrule
Base Model & $\text{65.91}_{-7.57}$ & $\text{36.53}_{-9.54}$  & $\text{6.55}_{-1.63}$  & $\text{36.91}_{-8.97}$  \\
RoP \cite{mu2025robustness} & $\text{65.91}_{-7.57}$ & $\text{35.30}_{-10.77}$  & $\text{6.01}_{-2.17}$  & $\text{34.25}_{-11.63}$   \\
SD \cite{agrawal2025enhancing} & \textbf{$\text{69.70}_{-3.78}$} & $\text{40.14}_{-5.93}$  & $\text{7.06}_{-1.12}$  & \textbf{$\text{42.16}_{-3.72}$ }   \\
VTI \cite{liu2024reducing} & $\text{61.36}_{-12.12}$ & $\text{34.29}_{-11.78}$  & $\text{6.12}_{-2.06}$  & $\text{34.88}_{-11.00}$   \\
Ours & $\text{67.92}_{-5.56}$ & \textbf{$\text{41.10}_{-4.97}$} & \textbf{$\text{7.31}_{-0.87}$ } & $\text{39.61}_{-6.27}$    \\
\midrule
\midrule

\multicolumn{5}{c}{X-ray Original VQA}\\
\midrule
Base Model & 88.89 & 53.68 & 7.54 & 40.41   \\
\midrule

\multicolumn{5}{c}{X-ray Patient Movement}\\
\midrule
Base Model & $\text{80.56}_{-8.33}$ & $\text{47.83}_{-5.85}$  & $\text{6.40}_{-1.14}$  & $\text{34.59}_{-5.82}$  \\
RoP \cite{mu2025robustness} & $\text{63.89}_{-25.00}$ & $\text{42.23}_{-11.45}$  & $\text{5.53}_{-2.01}$  & $\text{32.07}_{-8.34}$   \\
SD \cite{agrawal2025enhancing} & $\text{73.61}_{-15.28}$ & $\text{45.66}_{-8.02}$  & $\text{6.37}_{-1.17}$  & $\text{35.73}_{-4.68}$    \\
VTI \cite{liu2024reducing} &  $\text{79.17}_{-9.72}$ & $\text{50.77}_{-2.91}$  & $\text{7.10}_{-0.44}$  & $\text{38.89}_{-1.52}$    \\
Ours & \textbf{$\text{87.50}_{-1.39}$} & \textbf{$\text{51.99}_{-1.69}$}  & \textbf{$\text{7.24}_{-0.30}$}  & \textbf{$\text{39.67}_{-0.74}$}   \\

\bottomrule
\end{tabular}}
\end{table}

\subsection{Implementation Details}
We construct two perturbed datasets with VQA pairs derived from SLAKE \cite{liu2021slake} and OmniMed \cite{hu2024omnimedvqa}, and we present the experimental results on SLAKE-based dataset in the main paper, while results on the OmniMed-based dataset are provided in the supplementary material.
We use Qwen2.5-VL-7B \cite{bai2025qwen2} as our base model for all experiments in this section.
In this paper, the number of selected sample pairs for building the prototype pool is 100 and the number of clusters is 8.
The number of parallel micro loops $k$ is set to 10 and the number of macro loop rounds $n$ is set to 2.
The calibration weight $\alpha$ is set to 0.05.
All experiments were carried out on 2 NVIDIA RTX A6000 GPUs (48 GB). 
We evaluate model performance using Accuracy on closed-ended questions and ROUGE-1, BLEU, and Recall on open-ended questions.
We compare our IMC method with state-of-the-art methods: 1) Robustness of Prompting (RoP) \cite{mu2025robustness} employs few-shot prompting to enhance robustness; 2) Self-Denoising (SD) \cite{agrawal2025enhancing} adopts prompt engineering; and 3)Visual and Textual Intervention (VTI) \cite{liu2024reducing} utilizes embedding rectifications for hallucination reduction.

\subsection{Performance on Image Noise} 
Table \ref{tab:img_pert} shows that the MLLM suffers performance drop under six image noise types across three medical modalities, confirming insufficient robustness against image perturbations.
Specifically, on open-ended questions, the base model accuracy drops by 6.56\%, 22.95\%, and 1.64\% for MRI (human motion, aliasing, banding), 13.63\% and 7.57\% for CT (sparse view, low dose), and 8.33\% for X-ray patient movement.
Meanwhile, on closed-ended questions, the ROUGE scores degrade across all noise types, with drops ranging from 2.09\% (MRI banding) to 27.59\% (MRI aliasing).
Notably, the base MLLM suffers particularly severe deterioration on MRI aliasing, as aliasing creates overlapping artifacts that destroy the overall image structure, unlike other noise types that only affect details and resolution.

\textbf{Comparison Methods.} The comparison methods RoP and SD exhibit improvements only in specific scenarios (12.67\%/9.65\% higher ROUGE scores under MRI aliasing and 3.79\%/2.27\% ACC increases under CT sparse view), indicating their substantial limitations and poor generalizability to diverse noise types.
Similarly, VTI shows unstable performance, achieving 4.3\% higher recall on X-ray patient movement but deteriorating on CT noise types, even performing worse than the original model.
This occurs because it employs only a single direction calibration, which is inadequate for diverse scenarios and may adversely pull the vision features further away from the correct feature space.

\textbf{Our Approach.} 
Compared to other methods, our proposed IMC framework is more robust and stable, bringing consistent performance improvement across all noise types.
Notably, our algorithm attains a substantial 13.28\% ROUGE score enhancement on open-ended questions under MRI aliasing and achieves 6.94\% accuracy improvement on closed-ended questions under X-ray patient movement.
This experiment indicates that our IMC framework can effectively enhance the robustness of MLLMs against various image artifacts, 
laying a solid foundation for the reliability of medical MLLMs in practical clinical applications.

\subsection{Performance on Text Noise}
Similar to vision artifacts, Table \ref{tab:text_pert} demonstrates the sensitivity of the base MLLM to five noise types, resulting in clear performance drops.
Specifically, the model exhibits consistent sensitivity patterns across diverse character perturbations, with closed-ended question accuracy declining by 1.84\% to 5.35\% and open-ended question metrics showing average reductions of approximately 4.5\% for ROUGE and 3\% for BLEU.
The MLLM shows more severe accuracy deterioration under unrelated sentence noise, experiencing a 19.97\% performance decline on closed-ended questions and 10.03\% ROUGE drops on open-ended questions.
Moreover, the degradation variance from these character-level modifications is smaller than from vision noise, likely due to the comparable perturbation severity that fall beyond the textual tokenizer's effective encoding range.

\textbf{Comparison Methods.} The comparison methods RoP and SD cannot effectively reduce textual noise influence, as they lack the ability to identify textual corruptions and the noise contaminates the entire processing context.
Meanwhile, VTI performs more inconsistently on text noise compared to vision corruptions, yielding several negative effects across all noise types. 
This may result from the inadequacy of latent space steering in calibrating perturbations, particularly when noisy words (e.g., ``these" becoming ``th" and ``ese") induce out-of-vocabulary tokens for the tokenizer.

\textbf{Our Approach.} Conversely, our approach demonstrates effective robustness improvement by creating iterative multi-agent system. 
Specifically, under random character deletion noise, we achieve 2.41\% ROUGE improvement on open-ended questions and 2.67\% accuracy increase on closed-ended questions. 
The improvements become even more pronounced under unrelated sentences perturbation, reaching 5.03\% and 12.73\% respectively.
This experiment strongly supports that our IMC framework can effectively reduce MLLMs' sensitivity to different textual noise types.

\begin{table}[h]
\centering
\fontsize{8pt}{8pt}\selectfont
\addtocounter{table}{0}
\renewcommand{\arraystretch}{1.1}
\setlength{\tabcolsep}{1.2mm}
{
\caption{Performance comparison of MLLM robustness across five text perturbation types, subscripts denote performance drops. \textbf{Bold} font denotes the best performance.}
\label{tab:text_pert}
\begin{tabular}{>{\centering\arraybackslash}c|cccc}
\toprule
Method & $\textrm{ACC($\uparrow$)}$ & $\textrm{ROUGE($\uparrow$)}$  & $\textrm{BLEU($\uparrow$)}$  & $\textrm{Recall($\uparrow$)}$ \\
\midrule
\midrule
\multicolumn{5}{c}{Original VQA}\\
\midrule
Base Model & 75.09 & 49.71 & 7.93 & 44.96   \\
\midrule

\multicolumn{5}{c}{Delete Characters}\\
\midrule
Base Model & $\text{70.58}_{-4.51}$ & $\text{44.17}_{-5.54}$  & $\text{7.08}_{-0.85}$  & $\text{40.70}_{-4.26}$    \\
RoP \cite{mu2025robustness}  & $\text{68.82}_{-6.27}$ & $\text{37.04}_{-12.67}$  & $\text{5.91}_{-2.02}$  & $\text{36.39}_{-8.57}$   \\
SD \cite{agrawal2025enhancing} & $\text{73.06}_{-2.03}$ & $\text{39.38}_{-10.33}$  & $\text{6.18}_{-1.75}$  & $\text{37.53}_{-7.43}$    \\
VTI \cite{liu2024reducing} & $\text{70.10}_{-4.99}$ & $\text{42.16}_{-7.55}$  & $\text{6.84}_{-1.09}$  & $\text{39.17}_{-5.79}$  \\
Ours & \textbf{$\text{73.25}_{-1.84}$} & \textbf{$\text{46.58}_{-3.13}$}  & \textbf{$\text{7.45}_{-0.48}$}  & \textbf{$\text{42.39}_{-2.57}$}   \\
\midrule

\multicolumn{5}{c}{Add Characters}\\
\midrule
Base Model & $\text{73.25}_{-1.84}$ & $\text{45.32}_{-4.39}$  & $\text{7.34}_{-0.59}$  & $\text{42.19}_{-2.77}$    \\
RoP \cite{mu2025robustness} & \textbf{$\text{73.80}_{-1.29}$} & $\text{38.94}_{-10.77}$  & $\text{5.97}_{-1.96}$  & $\text{38.21}_{-6.75}$   \\
SD \cite{agrawal2025enhancing} & $\text{72.51}_{-2.58}$ & $\text{38.38}_{-11.33}$  & $\text{6.01}_{-1.92}$  & $\text{37.32}_{-7.64}$    \\
VTI \cite{liu2024reducing} & $\text{72.76}_{-2.33}$ & $\text{45.59}_{-4.12}$  & $\text{7.21}_{-0.72}$  & $\text{41.44}_{-3.52}$   \\
Ours & $\text{73.43}_{-1.66}$ & \textbf{$\text{47.45}_{-2.26}$}  & \textbf{$\text{7.77}_{-0.16}$}  & \textbf{$\text{43.41}_{-1.55}$}  \\
\midrule

\multicolumn{5}{c}{Swap Characters}\\
\midrule
Base Model & $\text{69.74}_{-5.35}$ & $\text{44.42}_{-5.29}$  & $\text{7.19}_{-0.74}$  & $\text{41.26}_{-3.70}$    \\
RoP \cite{mu2025robustness} & $\text{62.55}_{-12.54}$ & $\text{35.32}_{-14.39}$  & $\text{5.44}_{-2.49}$  & $\text{33.63}_{-11.33}$   \\
SD \cite{agrawal2025enhancing} & $\text{68.27}_{-6.82}$ & $\text{37.73}_{-11.98}$  & $\text{5.91}_{-2.02}$  & $\text{36.40}_{-8.56}$    \\
VTI \cite{liu2024reducing} & $\text{67.81}_{-7.28}$ & $\text{43.88}_{-5.83}$  & $\text{7.11}_{-0.82}$  & $\text{40.69}_{-4.27}$   \\
Ours & \textbf{$\text{69.93}_{-5.16}$} & \textbf{$\text{45.30}_{-4.41}$}  & \textbf{$\text{7.32}_{-0.61}$}  & \textbf{$\text{41.03}_{-3.93}$}    \\
\midrule

\multicolumn{5}{c}{Replace Characters}\\
\midrule
Base Model & $\text{70.48}_{-4.61}$ & $\text{45.03}_{-4.68}$  & $\text{7.18}_{-0.75}$  & $\text{41.23}_{-3.73}$    \\
RoP \cite{mu2025robustness} & $\text{65.87}_{-9.22}$ & $\text{35.20}_{-14.51}$  & $\text{5.46}_{-2.47}$  & $\text{34.64}_{-10.32}$   \\
SD \cite{agrawal2025enhancing} & $\text{70.11}_{-4.98}$ & $\text{38.45}_{-11.26}$  & $\text{5.94}_{-1.99}$  & $\text{36.99}_{-7.97}$    \\
VTI \cite{liu2024reducing} & $\text{70.86}_{-4.23}$ & $\text{43.81}_{-5.90}$  & $\text{7.16}_{-0.77}$  & $\text{40.47}_{-4.49}$   \\
Ours & \textbf{$\text{71.59}_{-3.50}$} & \textbf{$\text{46.96}_{-2.75}$}  & \textbf{$\text{7.62}_{-0.31}$}  & \textbf{$\text{42.64}_{-2.32}$}  \\
\midrule

\multicolumn{5}{c}{Unrelated Sentences}\\
\midrule
Base Model & $\text{58.12}_{-16.97}$ & $\text{39.68}_{-10.03}$  & $\text{6.38}_{-1.55}$  & $\text{37.00}_{-7.96}$   \\
RoP \cite{mu2025robustness} & $\text{54.23}_{-20.86}$ & $\text{32.74}_{-16.97}$  & $\text{5.56}_{-2.37}$  & $\text{34.83}_{-10.13}$   \\
SD \cite{agrawal2025enhancing} & $\text{64.22}_{-10.87}$ & $\text{36.84}_{-12.87}$  & $\text{6.38}_{-1.55}$  & $\text{38.86}_{-6.10}$    \\
VTI \cite{liu2024reducing} & $\text{67.62}_{-7.47}$ & $\text{41.73}_{-7.98}$  & $\text{6.81}_{-1.12}$  & $\text{38.14}_{-6.82}$   \\
Ours & \textbf{$\text{70.85}_{-4.24}$} & \textbf{$\text{44.71}_{-4.99}$}  & \textbf{$\text{7.17}_{-0.76}$}  & \textbf{$\text{40.77}_{-4.19}$}    \\

\bottomrule
\end{tabular}}
\end{table}

\subsection{Ablation Study}
To further validate our approach, we investigate the impact of layer-wise prototype numbers for image denoising and the number of macro loop iterations for textual processing.

\subsubsection{Prototype Numbers for Image Denoising}
We hypothesize that a larger number of prototypes yields finer-grained calibration direction vectors, thereby enhancing the precision of noisy embedding rectification.
This ablation study explores the impact of varying prototype numbers (1, 2, 4, 8, and 16) on denoising effectiveness under CT low dose setting with 100 sample pairs for building prototype pool.
As shown in Table \ref{tab:cluster_center}, ROUGE scores on open-ended questions gradually improve as the number of clusters increases, reaching the highest score of 41.10\% at 8 prototypes, while further increasing to 16 prototypes leads to performance degradation.
This phenomenon arises from the fact that finer-grained feature distribution partitioning provides more precise noise removal.
However, given the constraint of limited sample pairs, excessive division could introduce additional biases, resulting in performance drop.

\begin{table}[h]
\centering
\fontsize{9pt}{9pt}\selectfont
\addtocounter{table}{0}
\renewcommand{\arraystretch}{1.1}
\setlength{\tabcolsep}{1.8mm}
{
\caption{Ablation on prototype numbers under CT low dose noise.}
\label{tab:cluster_center}
\begin{tabular}{>{\centering\arraybackslash}c|cccc}
\toprule
Clusters & ACC($\uparrow$) & ROUGE($\uparrow$) & BLEU($\uparrow$) & Recall($\uparrow$) \\
\midrule
1 & 61.36 & 34.29 & 6.12 & 34.88 \\
2 & 65.15 & 36.59 & 6.63 & 37.53 \\
4 & 62.88 & 37.38 & 6.48 & 37.80 \\
8 & 67.92 & 41.10 & 7.31 & 39.61 \\
16 & 69.70 & 37.80 & 6.82 & 37.42 \\
\bottomrule
\end{tabular}}
\end{table}

\subsubsection{Macro Loop Rounds for Textual Denoising}
For textual noise processing in our IMC framework, the number of macro loop rounds can directly control the final denoising quality, but it also increases the time complexity.
To optimize this trade-off, we explore the relationship between macro loop rounds and denoising quality, evaluating configurations with 1 to 4 loop iterations. 
We perform experiments using character level random swap noise to facilitate fair comparison.
As demonstrated in Table \ref{tab:loop_rounds}, while ROUGE, BLEU, and Recall scores on open-ended questions increase with loop rounds, but the marginal improvements become negligible after 2 rounds.
In order to balance the time complexity with performance, and considering the minimal improvement on close-ended questions, we set the loop rounds to 2 in our implementation.

\begin{table}[h]
\centering
\fontsize{9pt}{9pt}\selectfont
\addtocounter{table}{0}
\renewcommand{\arraystretch}{1.1}
\setlength{\tabcolsep}{1.6mm}
{
\caption{Macro loop rounds ablation under character swap noise.}
\label{tab:loop_rounds}
\begin{tabular}{>{\centering\arraybackslash}c|cccc}
\toprule
Loop Rounds & ACC($\uparrow$) & ROUGE($\uparrow$) & BLEU($\uparrow$) & Recall($\uparrow$) \\
\midrule
1 & 68.27 & 41.86 & 6.79 & 38.28 \\
2 & 69.93 & 45.30 & 7.32 & 41.03 \\
3 & 69.56 & 45.32 & 7.32& 41.43 \\
4 & 68.63 & 46.05 & 7.48 & 42.19 \\
\bottomrule
\end{tabular}}
\end{table}

%% file: sec/6_con.tex
\section{Conclusion}
\label{sec:Con}
In this paper, we introduce the RobustMed-Bench that is designed to simulate various noise types in medical multi-modal data. 
Our experimental analysis shows that MLLMs exhibit significant performance degradation across different modalities when exposed to input perturbations.
To address this issue, we develop a training-free framework called Inherent-enhanced Multi-modal Calibration (IMC) consisting of Perturbation-aware Denoising Calibration (PDC) and Self-instantiated Multi-agent System (SMS), which leverages MLLMs' inherent capability through a perceive-and-calibrate paradigm for multi-modal noise removal.
This study represents the pioneering effort in benchmarking and enhancing the robustness of medical MLLMs, significantly improving their practical clinical applicability.

%% file: sec/X_suppl.tex
\maketitlesupplementary

\section{OmniMed-based Benchmark}
\label{sec:Benchmark}

\begin{figure}[h]
  \centering
   \includegraphics[width=1\linewidth]{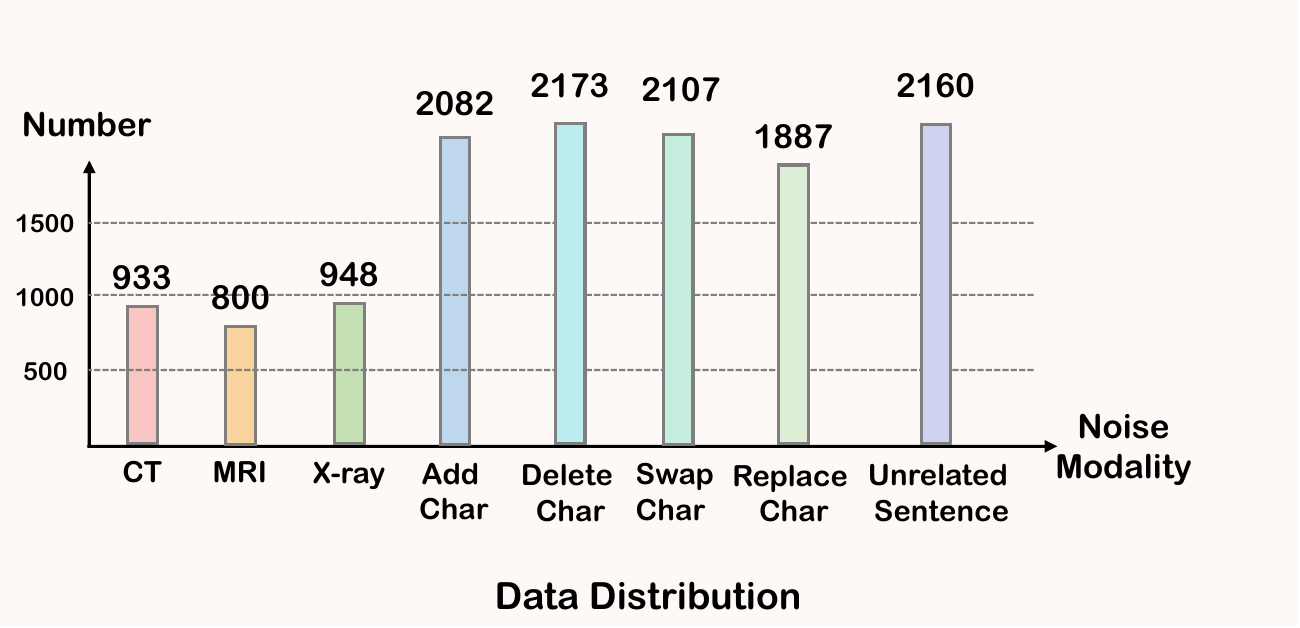}
  \caption{Statistical distribution of modalities and question categories of the OmniMed-based benchmark.
  }
  \label{fig:omni-based}
\end{figure}

We construct another benchmark based on OmniMed \cite{hu2024omnimedvqa} due to its data diversity, which encompasses multiple medical imaging modalities.
Similar to the SLAKE-based benchmark mentioned in our main paper, we also construct six different image artifacts across three imaging modalities and 5 different types of language noise, including character-level and sentence-level perturbations.
Figure \ref{fig:omni-based} shows the data distribution of the OmniMed-based benchmark, where we select 933/800/948 samples for CT/MRI/X-ray modalities, respectively.
It is worth noting that the OmniMed-based benchmark consists exclusively of Multiple-Choice Questions (MCQ), with each question providing four different choices. 
Therefore, for evaluating model performance on this benchmark, we simply report the accuracy.

\section{Comprehensive Results on SLAKE-based Benchmark}
On the SLAKE-based benchmark, we report the performance degradation of two closed-source MLLMs (GPT-5-Mini \cite{openai2025gpt5}, GPT-4o-Mini \cite{hurst2024gpt}) and two closed-source MLLMs (InternVL-3-9B \cite{zhu2025internvl3} and Lingshu-7B \cite{xu2025lingshu}).
As demonstrated in Table \ref{tab:exp_slake_close}, we notice that MRI aliasing artifacts bring the most severe impact to model performance compared to other imaging noise, which aligns with our observations mentioned in the main paper. 
The performance degradation under MRI aliasing reaches 26.25\%/15.57\%/17.21\%/35.24\% accuracy reduction on closed-ended questions and 21.99\%/15.25\%/11.47\%/22.54\% ROUGE score drops on open-ended questions for GPT-5-Mini/GPT-4o-Mini/InternVL/Lingshu, respectively.
On the other hand, X-ray patient movement has minimal impact with no obvious performance decreases on both open-ended and closed-ended questions across the three open-source MLLMs.
This phenomenon may arise because X-ray images have higher modal distinctiveness, and minor patient movements cannot significantly distort the essential anatomical structures needed for organ identification and modality classification.
These experimental results emphasize that noise in both image and text can affect model performance, which hampers their practical deployment in real-world clinical scenarios.

\begin{table*}[h]
\center
\caption{The experiments carried out on SLAKE-based benchmark, which shows the performance comparison of open-source and closed-source MLLMs robustness across six medical imaging artifacts and five textual noise types. Subscripts denote performance drops.}
 \renewcommand{\arraystretch}{1.4}
    \setlength{\tabcolsep}{2mm}
    \fontsize{8.5pt}{8.5pt}\selectfont{
    
\begin{tabular}{l|c|c|cccc}
\hline \hline
MLLMs & Modality & Noise & $\textrm{ACC($\uparrow$)}$ & $\textrm{ROUGE($\uparrow$)}$  & $\textrm{BLEU($\uparrow$)}$  & $\textrm{Recall($\uparrow$)}$  \\
\hline
\multirow{9}{*}{\raisebox{-0ex}{GPT-5-Mini \cite{openai2025gpt5}}} 
& \multirow{4}{*}{MRI}  & MRI Original 
    & 80.33 & 45.79 & 7.37 & 48.10 \\
\cline{3-7}
& & MRI Human Motion 
    & $\text{72.13}_{-8.20}$ 
    & $\text{45.53}_{-0.26}$  
    & $\text{7.32}_{-0.05}$  
    & $\text{46.12}_{-1.98}$ \\
& & MRI Aliasing 
    & $\text{54.10}_{-26.23}$ 
    & $\text{23.80}_{-21.99}$  
    & $\text{3.31}_{-4.06}$  
    & $\text{20.80}_{-27.30}$ \\
& & MRI Banding 
    & $\text{72.95}_{-7.38}$ 
    & $\text{45.11}_{-0.68}$  
    & $\text{7.54}_{+0.17}$  
    & $\text{45.33}_{-2.77}$ \\
\cline{3-7}

\cline{2-7}

& \multirow{3}{*}{CT}  & CT Original 
    & 79.64 & 49.23 & 7.68 & 45.18 \\
\cline{3-7}
& & CT Sparse View 
    & $\text{78.21}_{-1.43}$ 
    & $\text{42.79}_{-6.44}$  
    & $\text{6.19}_{-1.49}$  
    & $\text{36.89}_{-8.29}$ \\
& & CT Low Dose 
    & $\text{79.29}_{-0.35}$ 
    & $\text{48.72}_{-0.51}$  
    & $\text{7.54}_{-0.14}$  
    & $\text{44.86}_{-0.32}$ \\

\cline{2-7}

& \multirow{2}{*}{X-ray}  & X-ray Original 
    & 68.57 & 46.98 & 6.68 & 40.49 \\
\cline{3-7}
& & X-ray Patient Movement 
    & $\text{66.43}_{-2.14}$ 
    & $\text{46.67}_{-0.31}$  
    & $\text{6.53}_{-0.15}$  
    & $\text{39.96}_{-0.53}$ \\

\hline
\hline

\multirow{9}{*}{\raisebox{-0ex}{GPT-4o-Mini \cite{hurst2024gpt}}} 
& \multirow{4}{*}{MRI}  & MRI Original 
    & 74.59 & 38.89 & 5.73 & 36.82 \\
\cline{3-7}
& & MRI Human Motion 
    & $\text{70.49}_{-4.10}$ 
    & $\text{28.17}_{-10.72}$  
    & $\text{4.41}_{-1.32}$  
    & $\text{26.98}_{-9.84}$ \\
& & MRI Aliasing 
    & $\text{59.02}_{-15.57}$ 
    & $\text{23.64}_{-15.25}$  
    & $\text{3.56}_{-2.17}$  
    & $\text{21.85}_{-14.97}$ \\
& & MRI Banding 
    & $\text{74.59}_{-0.00}$ 
    & $\text{35.50}_{-3.39}$  
    & $\text{5.22}_{-0.51}$  
    & $\text{33.80}_{-3.02}$ \\

\cline{2-7}

& \multirow{3}{*}{CT}  & CT Original 
    & 70.71 & 34.78 & 5.40 & 33.02 \\
\cline{3-7}
& & CT Sparse View 
    & $\text{65.71}_{-5.00}$ 
    & $\text{29.11}_{-5.67}$  
    & $\text{4.87}_{-0.53}$  
    & $\text{29.10}_{-3.92}$ \\
& & CT Low Dose 
    & $\text{67.86}_{-2.85}$ 
    & $\text{27.91}_{-6.87}$  
    & $\text{4.15}_{-1.25}$  
    & $\text{26.05}_{-6.97}$ \\

\cline{2-7}

& \multirow{2}{*}{X-ray}  & X-ray Original 
    & 77.86 & 47.33 & 5.41 & 33.83 \\
\cline{3-7}
& & X-ray Patient Movement 
    & $\text{72.14}_{-5.72}$ 
    & $\text{42.48}_{-4.85}$  
    & $\text{4.65}_{-0.76}$  
    & $\text{29.33}_{-4.50}$ \\

\hline\hline
    
\multirow{15}{*}{\raisebox{-0ex}{InternVL-3-9B \cite{zhu2025internvl3}}} 
& \multirow{5}{*}{MRI}  & MRI Original 
    & 77.05 & 43.09 & 7.37 & 44.26 \\
\cline{3-7}
& & MRI Human Motion 
    & $\text{71.31}_{-5.74}$ 
    & $\text{38.21}_{-4.88}$  
    & $\text{6.44}_{-0.93}$  
    & $\text{39.57}_{-4.69}$ \\
& & MRI Aliasing 
    & $\text{59.84}_{-17.21}$ 
    & $\text{31.62}_{-11.47}$  
    & $\text{4.97}_{-2.40}$  
    & $\text{31.69}_{-12.57}$ \\
& & MRI Banding 
    & $\text{74.59}_{-2.46}$ 
    & $\text{42.34}_{-0.75}$  
    & $\text{6.60}_{-0.77}$  
    & $\text{42.73}_{-1.53}$ \\
\cline{3-7}
& & Character Noise 
    & $\text{65.57}_{-11.48}$ 
    & $\text{33.33}_{-9.76}$  
    & $\text{5.54}_{-1.83}$  
    & $\text{37.02}_{-7.24}$ \\
& & Sentence Noise 
    & $\text{67.21}_{-9.84}$ 
    & $\text{29.70}_{-13.39}$  
    & $\text{4.34}_{-3.03}$  
    & $\text{33.10}_{-11.16}$ \\

\cline{2-7}

& \multirow{5}{*}{CT}  & CT Original 
    & 78.57 & 37.68 & 6.61 & 40.23 \\
\cline{3-7}
& & CT Sparse View 
    & $\text{67.86}_{-10.71}$ 
    & $\text{24.90}_{-12.78}$  
    & $\text{4.06}_{-2.55}$  
    & $\text{25.89}_{-14.34}$ \\
& & CT Low Dose 
    & $\text{71.43}_{-7.14}$ 
    & $\text{36.65}_{-1.03}$  
    & $\text{6.53}_{-0.08}$  
    & $\text{38.47}_{-1.76}$ \\
\cline{3-7}
& & Character Noise 
    & $\text{62.86}_{-15.71}$ 
    & $\text{24.42}_{-13.26}$  
    & $\text{3.90}_{-2.71}$  
    & $\text{28.50}_{-11.73}$ \\
& & Sentence Noise 
    & $\text{68.93}_{-9.64}$ 
    & $\text{24.00}_{-13.68}$  
    & $\text{3.64}_{-2.97}$  
    & $\text{27.06}_{-13.17}$ \\

\cline{2-7}

& \multirow{5}{*}{X-ray}  & X-ray Original 
    & 72.14 & 37.08 & 5.62 & 40.56 \\
\cline{3-7}
& & X-ray Patient Movement 
    & $\text{72.86}_{+0.72}$ 
    & $\text{35.25}_{-1.83}$  
    & $\text{4.91}_{-0.71}$  
    & $\text{39.17}_{-1.39}$ \\
\cline{3-7}
& & Character Noise 
    & $\text{65.71}_{-6.43}$ 
    & $\text{23.58}_{-13.50}$  
    & $\text{3.34}_{-2.28}$  
    & $\text{32.17}_{-8.39}$ \\
& & Sentence Noise 
    & $\text{61.43}_{-10.71}$ 
    & $\text{22.06}_{-15.02}$  
    & $\text{1.75}_{-3.87}$  
    & $\text{30.35}_{-10.21}$ \\

\hline
\hline

\multirow{15}{*}{\raisebox{-0ex}{Lingshu-7B \cite{xu2025lingshu}}} 
& \multirow{5}{*}{MRI}  & MRI Original 
    & 91.80 & 86.47 & 8.49 & 14.25 \\
\cline{3-7}
& & MRI Human Motion 
    & $\text{84.43}_{-7.37}$ 
    & $\text{82.02}_{-4.45}$  
    & $\text{8.35}_{-0.14}$  
    & $\text{14.77}_{+0.52}$ \\
& & MRI Aliasing 
    & $\text{56.56}_{-35.24}$ 
    & $\text{63.93}_{-22.54}$  
    & $\text{6.96}_{-1.53}$  
    & $\text{12.92}_{-1.33}$ \\
& & MRI Banding 
    & $\text{88.52}_{-3.28}$ 
    & $\text{83.40}_{-3.07}$  
    & $\text{7.10}_{-1.39}$  
    & $\text{12.65}_{-1.60}$ \\
\cline{3-7}
& & Character Noise 
    & $\text{76.23}_{-15.57}$ 
    & $\text{77.75}_{-8.72}$  
    & $\text{7.56}_{-0.93}$  
    & $\text{12.88}_{-1.37}$ \\
& & Sentence Noise 
    & $\text{81.97}_{-9.83}$ 
    & $\text{83.37}_{-3.10}$  
    & $\text{8.98}_{+0.49}$  
    & $\text{15.34}_{+1.09}$ \\

\cline{2-7}

& \multirow{5}{*}{CT}  & CT Original 
    & 78.57 & 75.10 & 3.26 & 10.87 \\
\cline{3-7}
& & CT Sparse View 
    & $\text{77.14}_{-1.43}$ 
    & $\text{59.78}_{-15.32}$  
    & $\text{2.33}_{-0.93}$  
    & $\text{7.95}_{-2.92}$ \\
& & CT Low Dose 
    & $\text{75.71}_{-2.86}$ 
    & $\text{70.56}_{-4.54}$  
    & $\text{2.81}_{-0.45}$  
    & $\text{9.27}_{-1.60}$ \\
\cline{3-7}
& & Character Noise 
    & $\text{69.29}_{-9.28}$ 
    & $\text{63.42}_{-11.68}$  
    & $\text{2.51}_{-0.75}$  
    & $\text{8.45}_{-2.42}$ \\
& & Sentence Noise 
    & $\text{74.64}_{-3.93}$ 
    & $\text{68.69}_{-6.41}$  
    & $\text{2.85}_{-0.41}$  
    & $\text{9.52}_{-1.35}$ \\

\cline{2-7}

& \multirow{5}{*}{X-ray}  & X-ray Original 
    & 82.14 & 82.17 & 4.19 & 11.86 \\
\cline{3-7}
& & X-ray Patient Movement 
    & $\text{82.14}_{+0.00}$ 
    & $\text{76.97}_{-5.20}$  
    & $\text{3.53}_{-0.66}$  
    & $\text{10.11}_{-1.75}$ \\
\cline{3-7}
& & Character Noise 
    & $\text{77.14}_{-5.00}$ 
    & $\text{66.29}_{-15.88}$  
    & $\text{3.33}_{-0.86}$  
    & $\text{11.20}_{-0.66}$ \\
& & Sentence Noise 
    & $\text{73.57}_{-8.57}$ 
    & $\text{75.37}_{-6.80}$  
    & $\text{3.43}_{-0.76}$  
    & $\text{11.32}_{-0.54}$ \\

\hline \hline
\end{tabular}}
\label{tab:exp_slake_close}
\end{table*}

\begin{table*}[h]
\center
\caption{The experiments carried out on OmniMed-based benchmark, which shows the performance comparison of the robustness of based model and our method across five textual noise types. Subscripts denote performance drops.}
 \renewcommand{\arraystretch}{1.5}
    \setlength{\tabcolsep}{1mm}
    \fontsize{8.5pt}{8.5pt}\selectfont{
    
\begin{tabular}{c|c|c|c|c|c}
\hline \hline

\multirow{2}{*}{Data \& Model}  & \multicolumn{5}{c}{Text Noise} \\
\cline{2-6}
& {Delete Characters} & {Add Characters} & {Swap Characters}  & {Replace Characters} & {Unrelated Sentences}  \\
\hline
Original Data with Base Model & $\text{63.41}$ & $\text{64.02}$ & $\text{62.98}$ & $\text{64.71}$ & $\text{63.70}$ \\
Noisy Data with Base Model & $\text{58.49}_{-4.92}$ & $\text{58.98}_{-5.04}$ & $\text{58.14}_{-4.84}$ & $\text{59.67}_{-5.04}$ & $\text{57.64}_{-6.06}$ \\
Noisy Data with Our Approach & $\text{62.63}_{-0.78}$ & $\text{63.59}_{-0.43}$ & $\text{61.89}_{-1.09}$ & $\text{63.70}_{-1.01}$ & $\text{63.15}_{-0.55}$ \\

\hline \hline
\end{tabular}}
\label{tab:exp_omni_text}
\end{table*}

\section{Improvement of IMC Framework on OmniMed-based Benchmark}

On the OmniMed-based benchmark, we compare the base MLLM with our proposed denoising framework. 
Similar to the setting in our main paper, we choose the Qwen2.5-VL-7B as our base model.
The experimental results on textual noise are presented in Table \ref{tab:exp_omni_text} while the results on image artifacts are presented across three tables: Table \ref{tab:exp_omni_mri} for MRI noise, Table \ref{tab:exp_omni_ct} for CT noise, Table \ref{tab:exp_omni_xray} for X-ray noise.
Specifically, Table \ref{tab:exp_omni_text} demonstrates that our IMC framework can improve prediction accuracy of the base model by 4.14\%/4.61\%/3.75\%/4.03\% under randomly delete/add/swap/replace characters, and increases accuracy 5.51\% on additional unrelated sentence noise.
Tables \ref{tab:exp_omni_mri}-\ref{tab:exp_omni_xray} demonstrate that our IMC framework consistently outperforms the base model across all imaging modalities. 
Specifically, we achieve accuracy improvements of 1.84\%/4.30\%/3.06\% for MRI artifacts (human motion/aliasing/banding), 2.47\%/0.99\% for CT artifacts (low dose/sparse view), and 2.35\% for X-ray patient movement, respectively.

\begin{table*}[h]
\center
\caption{The experiments carried out on OmniMed-based benchmark, which shows the performance comparison of the robustness of based model and our method across three MRI artifacts. Subscripts denote performance drops.}
 \renewcommand{\arraystretch}{1.5}
    \setlength{\tabcolsep}{2mm}
    \fontsize{8.5pt}{8.5pt}\selectfont{
    
\begin{tabular}{c|c|c|c|c|c|c}
\hline \hline

\multicolumn{7}{c}{MRI} \\
\hline
Original & \multicolumn{2}{c|}{Human Motion} & \multicolumn{2}{c|}{ Aliasing} & \multicolumn{2}{c}{ Banding}  \\
\hline
Base Model & Base Model & Ours  & Base Model & Ours  & Base Model & Ours \\
\hline
 $\text{73.62}$ & $\text{68.71}_{-4.91}$  &  $\text{70.55}_{-3.07}$ &  $\text{48.77}_{-24.85}$ &  $\text{53.07}_{-20.55}$   &  $\text{69.33}_{-4.29}$ &  $\text{72.39}_{-1.23}$   \\

\hline \hline
\end{tabular}}
\label{tab:exp_omni_mri}
\end{table*}

\begin{table}[h]
\center
\caption{The experiments carried out on OmniMed-based benchmark, which shows the performance comparison of the robustness of based model and our method across two CT artifacts. Subscripts denote performance drops.}
 \renewcommand{\arraystretch}{1.5}
    \setlength{\tabcolsep}{1.1mm}
    \fontsize{8.5pt}{8.5pt}\selectfont{
    
\begin{tabular}{c|c|c|c|c}
\hline \hline

\multicolumn{5}{c}{CT} \\
\hline
Original & \multicolumn{2}{c|}{Low Dose} & \multicolumn{2}{c}{Sparse View}\\
\hline
Base Model & Base Model & Ours  & Base Model & Ours  \\
\hline
 $\text{54.68}$ & $\text{43.10}_{-11.58}$  &  $\text{45.57}_{-9.11}$ &  $\text{47.78}_{-6.90}$ &  $\text{48.77}_{-5.91}$    \\

\hline \hline
\end{tabular}}
\label{tab:exp_omni_ct}
\end{table}

\begin{table}[h]
\center
\caption{The experiments carried out on OmniMed-based benchmark, which shows the performance comparison of the robustness of based model and our method on X-ray artifacts. Subscripts denote performance drops.}
 \renewcommand{\arraystretch}{1.5}
    \setlength{\tabcolsep}{2mm}
    \fontsize{8.5pt}{8.5pt}\selectfont{
    
\begin{tabular}{c|c|c}
\hline \hline

\multicolumn{3}{c}{X-ray} \\
\hline
Original & \multicolumn{2}{c}{Patient Movement}\\
\hline
Base Model & Base Model & Ours \\
\hline
 $\text{78.35}$ & $\text{72.24}_{-6.11}$  &  $\text{74.59}_{-3.76}$\\

\hline \hline
\end{tabular}}
\label{tab:exp_omni_xray}
\end{table}

\section{Case Study}
\label{sec:case_study}
\begin{figure}[h]
  \centering
   \includegraphics[width=1\linewidth]{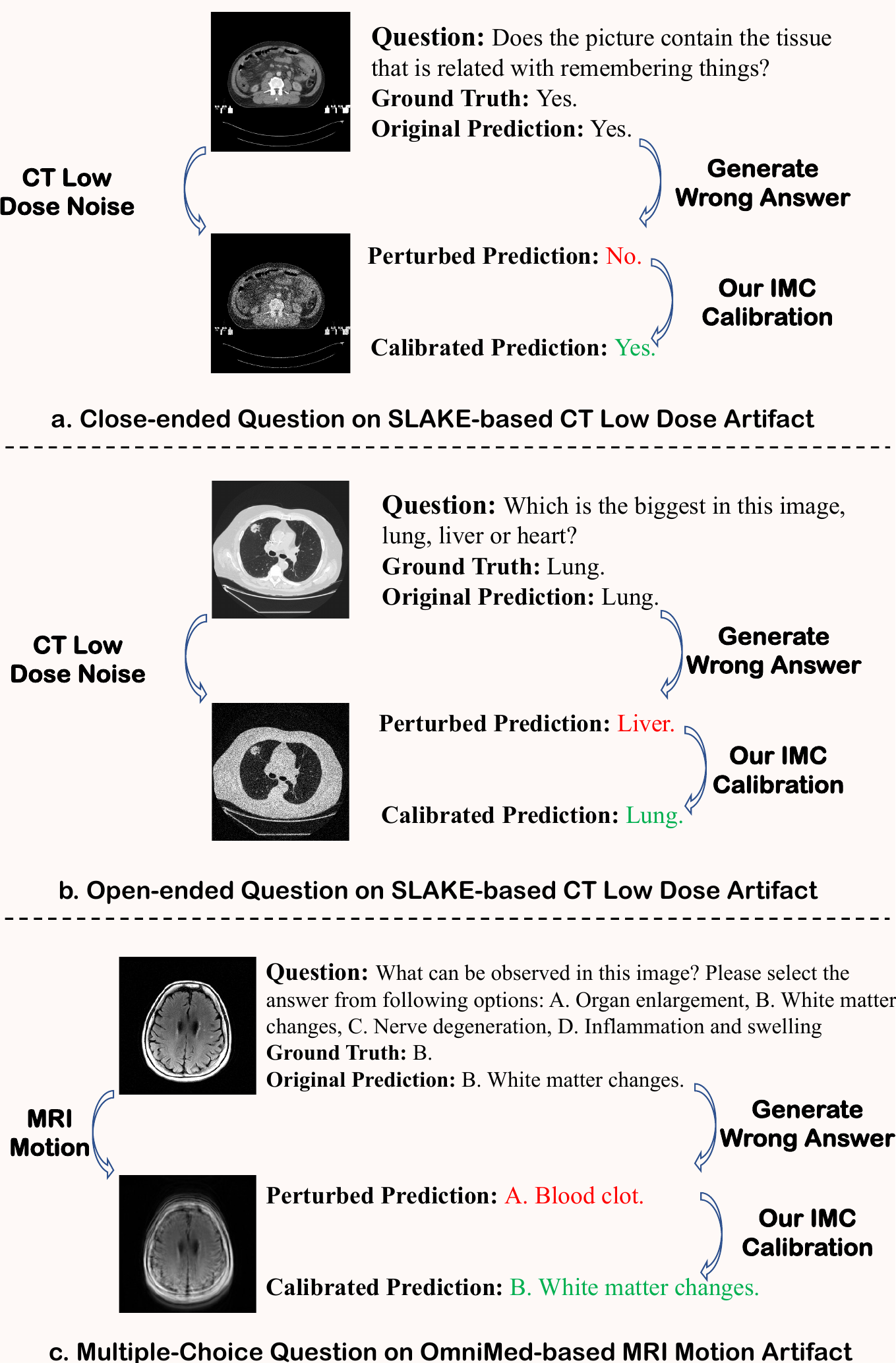}
  \caption{(a-b) Case studies of close-ended and open-ended questions on SLAKE-based CT low dose artifact; (c) Case study of Multiple-Choice Question on OmniMed-based MRI Motion noise.
  }
  \label{fig:case_low_dose}
\end{figure}

We visualize some cases in this section for demonstrating the effectiveness of our proposed denoising framework.
Figure \ref{fig:case_low_dose} presents the close-ended and open-ended questions on SLAKE-based benchmark perturbed by CT low dose artifact.
And the multiple-choice question on OmniMed-based MRI motion noise.
fact.
The base model initially generates correct responses. 
However, when encountering CT low-dose noise, the model produces incorrect results. 
After applying our method to calibrate the image latent features, the model is able to generate correct answers again.

\section{Prompts Used for Constructing Agents}
\begin{figure*}[t]
  \centering
   \includegraphics[width=1\linewidth]{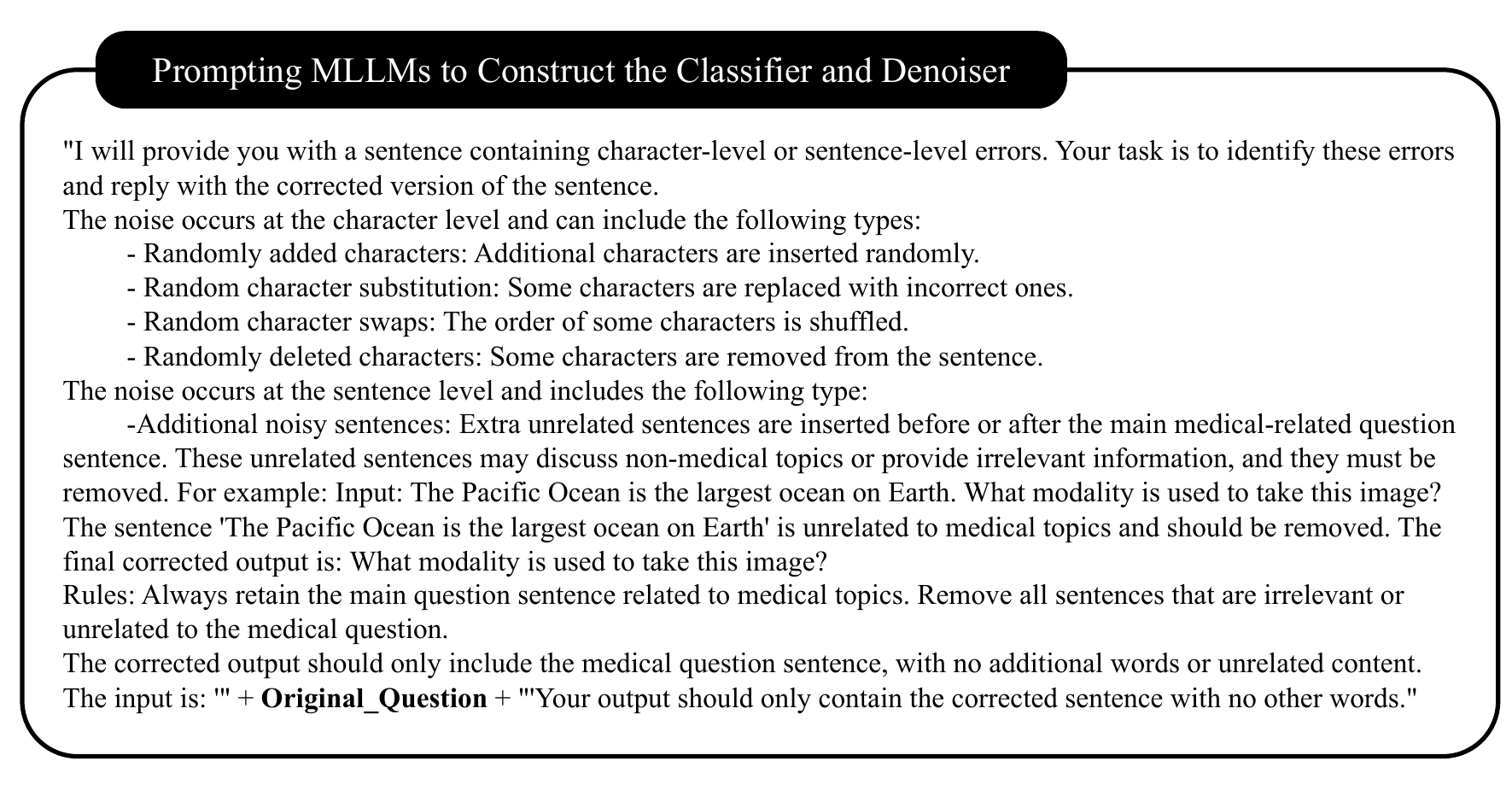}
  \caption{Message used to prompt the base MLLM to construct the first agent (i.e., \textit{Classifier and Denoiser}) that identifies noise types and performs corresponding noise removal operations. 
  The bold font \textbf{Original\_Question} represents the noisy question input for denoising.
  }
  \label{fig:agent1_prompt}
\end{figure*}

\begin{figure*}[h]
  \centering
   \includegraphics[width=1\linewidth]{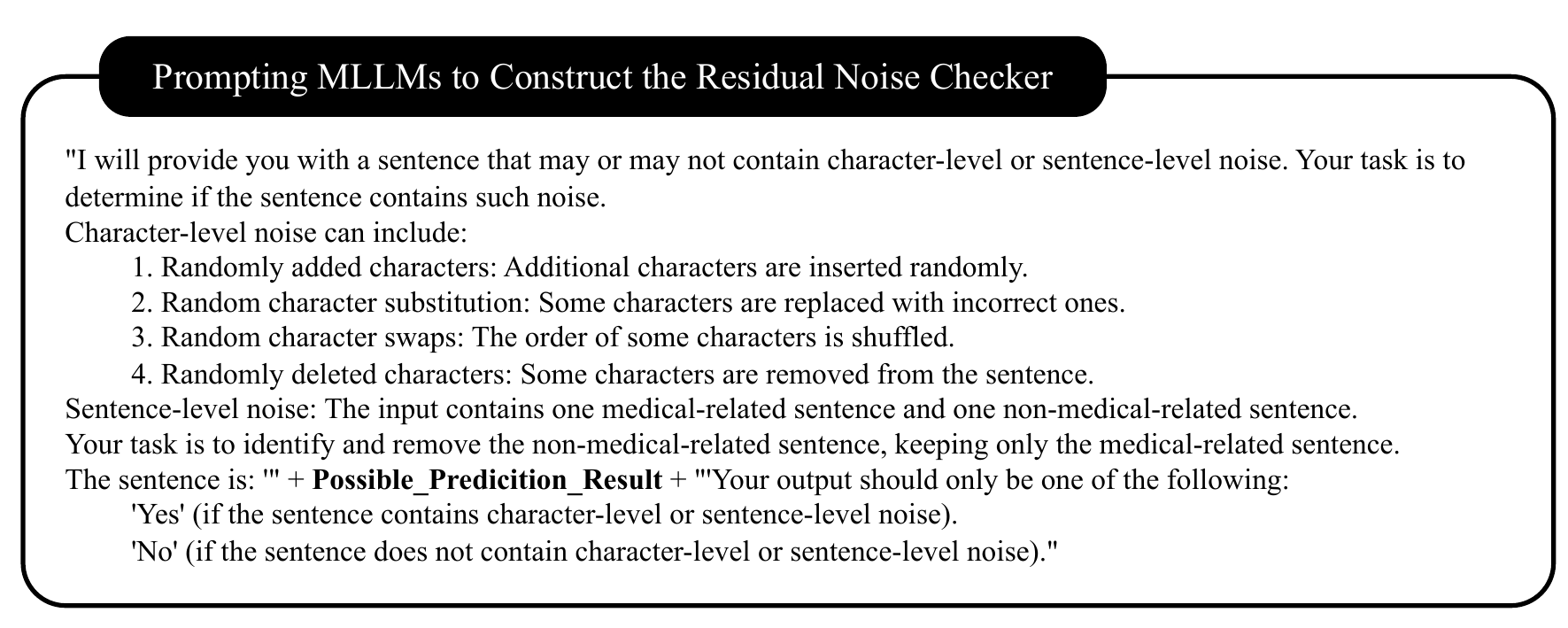}
  \caption{Message used to prompt the base MLLM to construct the second agent (i.e., \textit{Residual Noise Checker}) that determines whether the denoised results from previous step still contain noise. 
  The bold font \textbf{Possible\_Predicition\_Result} represents the denoised result outputed from the previous agent \textit{Classifier and Denoiser}.
  }
  \label{fig:agent2_prompt}
\end{figure*}

\begin{figure*}[h]
  \centering
   \includegraphics[width=1\linewidth]{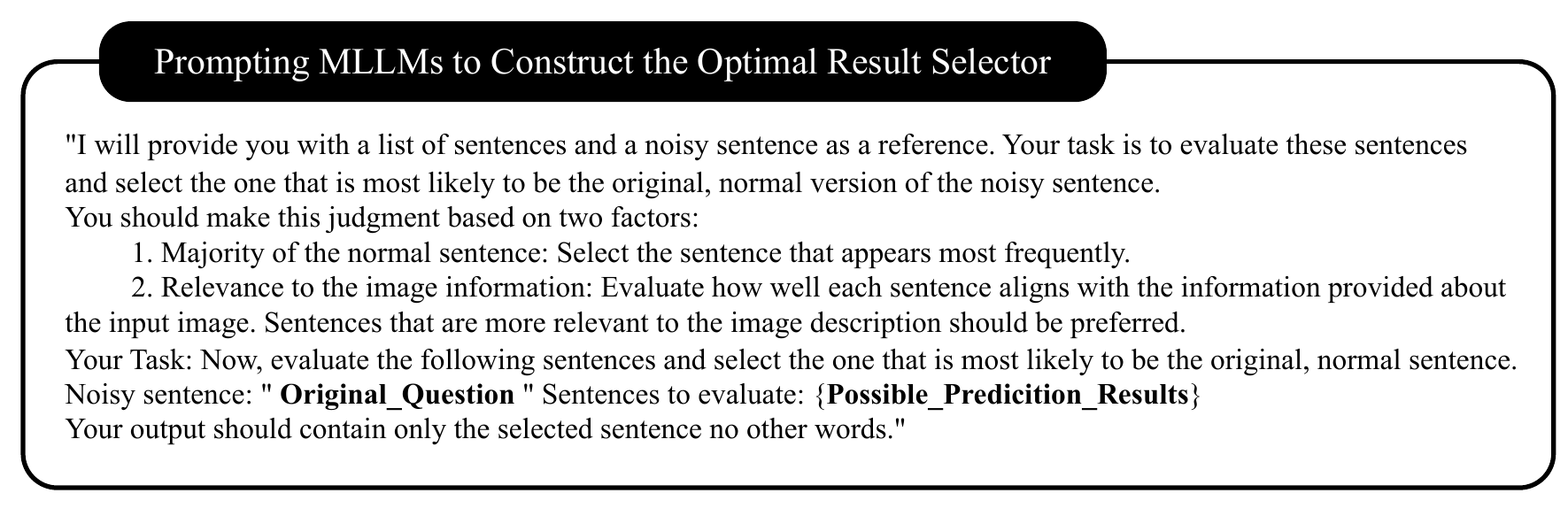}
  \caption{Message used to prompt the base MLLM to construct the third agent (i.e., \textit{Optimal Result Selector}). 
  This agent compares results from $k$ parallel micro loops to identify the output that is most noise-free compared to the input sentence and consistent with the visual information.
  The bold font \textbf{ Original\_Question} represents the input noisy sentence in this round, and \textbf{\{Possible\_Predicition\_Results\}} means the collection of $k$ outputs from previous micro loops.
  }
  \label{fig:agent3_prompt}
\end{figure*}

\begin{figure*}[h]
  \centering
   \includegraphics[width=1\linewidth]{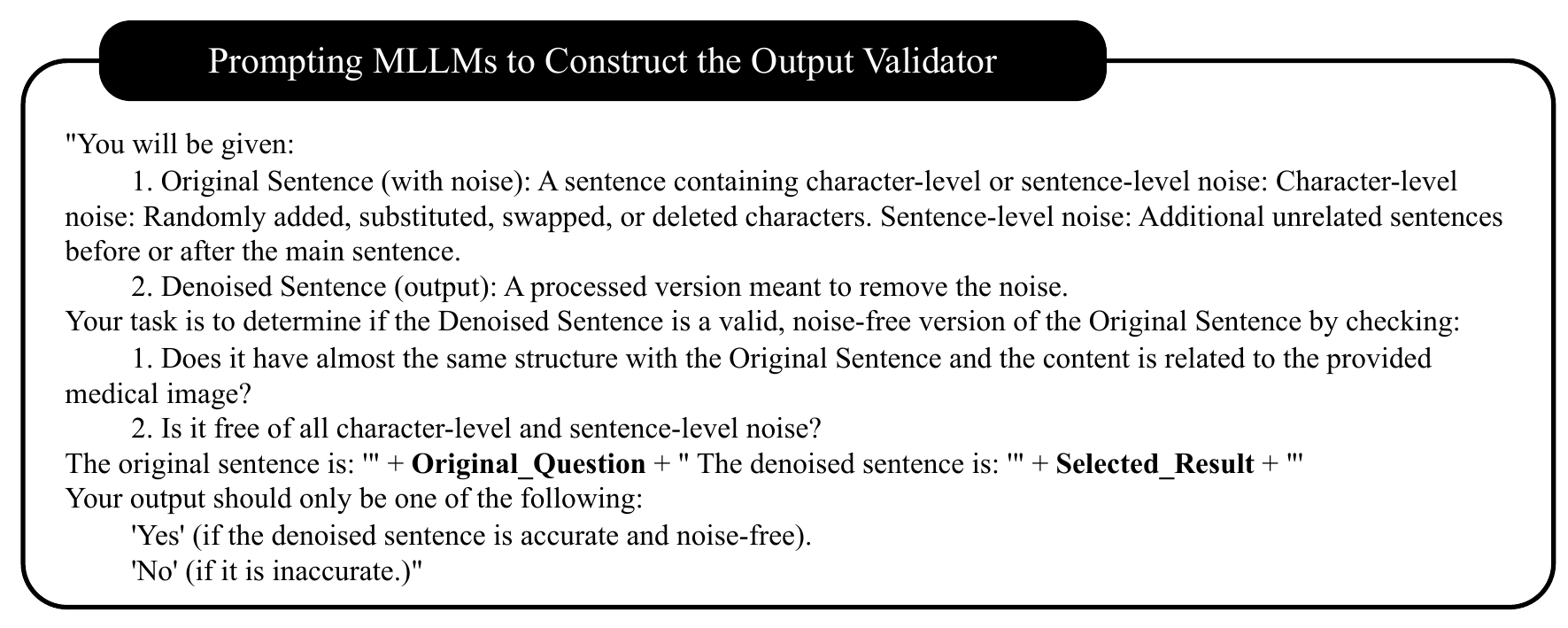}
  \caption{Message used to prompt the base MLLM to construct the last agent (i.e., \textit{Output Validator}) that determines whether the result is a coherent and logically consistent sentence.
  The bold font \textbf{ Original\_Question} represents the input noisy sentence in this round, and \textbf{\{Possible\_Predicition\_Results\}} means the collection of $k$ outputs from previous micro loops.
  }
  \label{fig:agent4_prompt}
\end{figure*}

In this section, we illustrate the prompts used for constructing agents for textual noise removal. Figure \ref{fig:agent1_prompt} shows the prompt for building the \textit{Classifier and Denoiser} that identifies and removes character-level and sentence-level noise. Figure \ref{fig:agent2_prompt} shows the prompt for the \textit{Residual Noise Checker} that verifies whether the denoised result still contains residual noise. Figure \ref{fig:agent3_prompt} shows the prompt for the \textit{Optimal Result Selector} that selects the most accurate denoised sentence from multiple candidates. Figure \ref{fig:agent4_prompt} shows the prompt for the \textit{Output Validator} that ensures the denoised sentence is noise-free and contextually consistent.

\label{sec:Benchmark}

